\documentclass{article} 
\usepackage{iclr2016_conference,times}
\usepackage{hyperref}
\usepackage{url}
\usepackage[utf8]{inputenc}
\usepackage{amsmath}
\usepackage{amsfonts}
\usepackage{amssymb}
\usepackage{graphicx}
\usepackage{caption}
\usepackage{subcaption}
\usepackage{dsfont}
\usepackage[toc,page]{appendix}

\newcommand{\argmin}{\operatornamewithlimits{argmin}}
\newcommand{\x}{\mathbf{x}}
\newcommand{\w}{\mathbf{w}}

\title{Auxiliary Image Regularization for Deep CNNs with Noisy Labels}

\author{Samaneh Azadi$^1$, Jiashi Feng$^{1,2}$, Stefanie Jegelka$^3$ \& Trevor Darrell$^1$ \\
$^1$ Department of EECS, University of California, Berkeley\\
$^2$ Department of ECE, National University of Singapore\\
$^3$ Department of EECS, Massachusetts Institute of Technology  \\
\texttt{\{sazadi,trevor\}@eecs.berkeley.edu} \\ 
\texttt{elefjia@nus.edu.sg} \\ 
\texttt{stefje@csail.mit.edu} 
}

\iclrfinalcopy 

\begin{document}

\maketitle
\begin{abstract}
Precisely-labeled data sets with sufficient amount of samples are very important for training  deep convolutional neural networks (CNNs). However, many of the available real-world data sets contain erroneously labeled samples and those errors substantially hinder the learning of very accurate CNN models. In this work, we consider the problem of training a deep CNN model for image classification with mislabeled training samples \textendash{} an issue that is common in  real image data sets with tags supplied by amateur users. To solve this  problem, we propose an auxiliary image regularization technique, optimized by the stochastic Alternating Direction Method of Multipliers (ADMM) algorithm, that automatically exploits the mutual context information among training images and encourages the model to  select reliable images to robustify the  learning process.  Comprehensive experiments on  benchmark data sets clearly demonstrate our proposed regularized CNN model is resistant to  label noise in training data.
\end{abstract}

\section{Introduction}

Deep Convolutional Neural Network (CNN) models have seen great success in solving general object recognition problems~\citep{krizhevsky2012imagenet}. However, due to their extremely huge parameter space,  the performance of deep models relies heavily on the availability of  a sufficiently large number of training examples.  In practice, collecting  images as well as their  accurate annotations at a large scale is usually tedious and expensive. On the other hand, there are millions of freely available images with user-supplied tags that can be easily collected from the web. Being able to exploit this rich resource seems promising for learning a deep classification model. The labels of these web images, however, tend to be much more noisy, and hence challenging to learn from.

\begin{figure}[t!]
\centering
	\includegraphics[width=\textwidth]{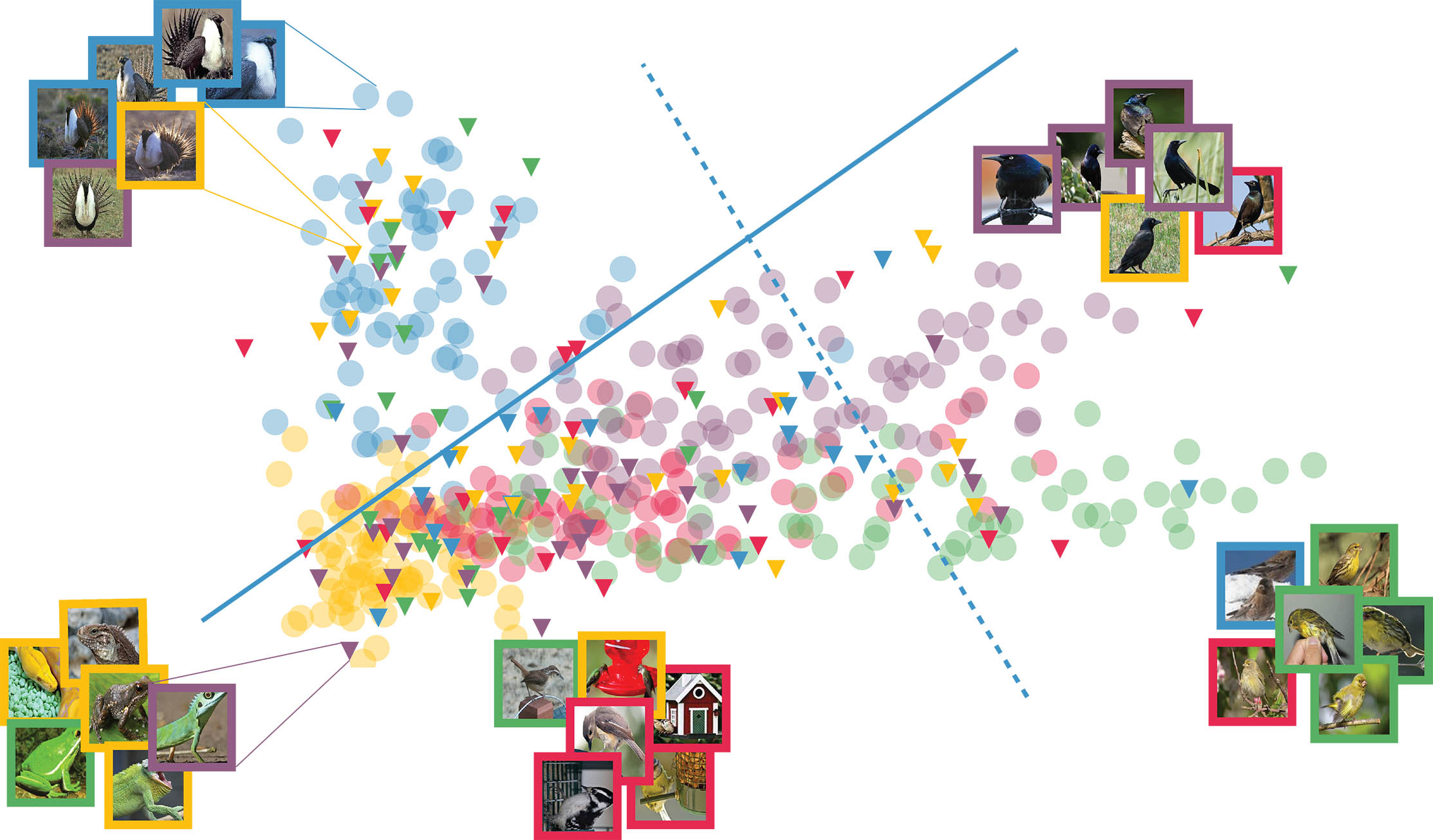}
    \caption{Exemplar description of the regularized deep CNN model. We impose the proposed auxiliary example regularization on the top layer of the deep CNN model. Image representations from 5 different categories of Imagenet7k data set, learned from a well-trained deep CNN model, are visualized by t-SNE embedding \citep{van2008visualizing}. The noisy categories are shown in different colors, circles demonstrate images with clean labels in the learned manifold and triangles display images with corrupted annotations. Example images are clustered according to their ground-truth labels. The border color indicates noise-contaminated labels of images. Our auxiliary image regularizer (AIR) gives higher weights to the important image representations while seeking for a structured sparsity pattern. AIR encourages the impact of some images to go to zero (especially those mislabeled images whose representations associate them with a set of images incorporated with a different label). It in fact pursues some ``nearest neighbors'' within the training examples to regularize the fitting of a deep CNN model to noisy samples. As the result, the groups (and the corresponding features) deemed not
relevant will have small effect on classification. The blue solid line is the hypothetical boundary classifier suggested by AIR to classify images in the blue category versus non-blue categories while the dashed line corresponds to the hypothetical SVM classifier. Here, AIR disregards images with blue noisy labels (triangles) while SVM considers those examples in training the classifier.}   
 \label{fig.models}
\end{figure}

In this work, we consider the problem of image classification in this challenging scenario where annotations of the training examples are noisy. In particular, we are interested in how to learn a deep CNN model  that is able to produce robust image representations and classification results in presence of  noisy supervision. Deep CNN models have been recognized to be sensitive to  sample and label noise in recent works \citep{sukhbaatar2014learning,nguyen2014deep}. Thus, several methods \citep{sukhbaatar2014learning,xiao2015learning}  have  been developed to alleviate the negative effect of noisy labels in learning the deep models.  However, most of the existing methods only consider modeling a fixed category-level label confusion distribution and cannot alleviate the effect of noise in representation learning for each sample.  

We propose a novel auxiliary image regularizer (AIR) to address this issue of deception of training annotations. Intuitively, the proposed regularizer exploits the structure of the data and automatically retrieves useful auxiliary examples to collaboratively facilitate training of the classification model. Here, structure of the data means the nonlinear manifold structure underlying images from multiple categories learned from a well-trained deep model on another data set $D^\prime$. To some extent, the AIR regularizer can be deemed as seeking some ``nearest neighbors'' within the  training examples  to regularize the fitting of a deep CNN model to noisy samples and improve its classification  performance in presence of noise.

Robustifying classification models  via regularization is a common practice to enhance  robustness of the models. Popular regularizers include Tikhonov regularization ($\ell_2$-norm) and the $\ell_1$-norm on the model parameters \citep{tibshirani1996regression}. However, an effective regularizer for training a deep CNN model is still absent, especially for handling the above learning problem with faulty labels. To the best of our knowledge, this work is among the first to introduce an effective regularizer for deep CNN models to handle label noise.

Inspired by related work in natural language processing \citep{yogatama2014making},
we use a group sparse norm to automatically select  auxiliary images. 
Figure~\ref{fig.models} shows an exemplar overview of our model. In contrast to previous works imposing the regularization on the model parameter, we propose to construct groups of input image features and apply the group sparse regularizer on the response maps. Imposing such group sparsity regularization on the classifier response enables it to actively select the relevant and useful features, which gives higher learning weights to the informative groups in the classification task and forces the weights of irrelevant or noisy groups toward zero. The activated auxiliary images implicitly provide guiding information for training the deep models. We solve the associated optimization problem via ADMM~\citep{glowinski1975approximation,gabay1976dual,lions79}, recently popularized by~\cite{boyd2011distributed}. In particular, we use the stochastic ADMM method of~\cite{ouyang2013stochastic} and~\cite{azadi2014sadmm} on this large-scale problem.

We demonstrate the effect of AIR on image classification via deep CNNs, where we synthetically corrupt the training annotations. We investigate how the proposed method identifies informative images and filters out noisy ones among the candidate auxiliary images. Going one step further, we then explore how the proposed method improves learning of image classification from user-supplied tags and handles the inherent noise in these tags.   
Comprehensive experiments on benchmark data sets, shown in Section~\ref{sec:expts}, clearly demonstrate the effectiveness of our proposed method for the large-scale image classification task.

\subsection{Related Work}

A large body of existing work proposes to employ sparsity-inducing \citep{tibshirani1996regression} or group-sparsity inducing \citep{friedman2010note,zou2005regularization} norms for effective model regularization and better model selection, with applications to dictionary learning \citep{mairal2009supervised} and image representation learning  \citep{yang2009linear,bach2012optimization}, to name a few examples from the field of computer vision. However, most focus on imposing a structured prior on the model parameters. The idea of exploiting the group sparsity structure within raw data  was recently proposed in \citep{yogatama2014making}, to solve  text recognition problems, by exploiting the intrinsic structure  among  sentences in a document. However, as far as we know, none of existing works  have investigated how to exploit the  structural information among data for  a deep  model, as we address here, and we are among the first to propose such a regularized deep model based on auxiliary data.

In this work, we are particularly interested in learning from noisy labeled image data, where a limited number of training examples are supplied with clean labels. Among the most recent contributions, \citet{sainbayar2015} solve this problem by learning the noise distribution through an extra noise layer added to the deep model,while \citet{xiao2015learning} explore this problem from a probabilistic graphical model point of view and train a classifier in an end-to-end learning procedure. \citet{izadinia2014image} introduce a robust logistic regression method for classification with user-supplied tags. \citet{feng2014robust} deal with arbitrary outliers in the data through a robust logistic regression method by estimating the parameters in a linear programming scheme. Different from those existing works, we automatically exploit the contextual information from useful auxiliary images via a new regularization on  deep CNN models. 

\section{Auxiliary Image Regularizer}

\subsection{Problem Setup }
\label{sec:problem}
Suppose we are given a  set of training images $ D= \{(\mathbf{x}_1,y_1), \ldots, (\mathbf{x}_n,y_n)\} \subset \mathbb{R}^{p} \times \mathcal{C} $. Some training samples in $D$ have noisy annotations. We do not assume any specific distribution on the noise in $y_i$. Indeed, we only assume the number of noisy labels does not exceed the number of correct labels. This significantly relaxes  the assumptions imposed in previous works \citep{sukhbaatar2014learning,xiao2015learning}. For instance, \citet{sukhbaatar2014learning} require the noise to follow a fixed confusion distribution. Our goal is to learn a deep CNN model $ \mathcal{M} $ for the data set $ D $ while we have access to a deep network pre-trained on an independent set $ D^\prime = \{(\mathbf{x}^\prime_1,y^\prime_1), \ldots, (\mathbf{x}^\prime_{n^\prime},y^\prime_{n^\prime})\}  \subset \mathbb{R}^{p} \times \mathcal{C}^\prime$ containing sufficiently many accurately annotated examples in categories $\mathcal{C}^\prime$. We use this pre-trained network to produce representations of images in the main set $D$. 
We employ the popular AlexNet architecture \citep{krizhevsky2012imagenet} to build our CNN model. The top layer which accounts for classification is parameterized by  $\mathbf{w}$, and we apply the usual empirical risk minimization method to learn such parameter: 
\begin{equation*}
\mathbf{w} = \argmin_{\w} \left\{L(\w;D) := \sum_{i=1}^n \ell(\mathbf{w};\mathbf{x}_i,y_i)\right\} ,
\end{equation*}
where $ \ell(\cdot;\mathbf{x}_i, y_i )$ is the classification loss on an individual training example in set $D$. Adding a regularizer to the loss function is a common approach in solving large-scale classification problems to prevent overfitting and train a more accurate and generalizable classifier:
\begin{eqnarray}
\label{eq.loss}
\hat{\w} = \argmin_{\w} L(\w;D)+\Omega(\w).
\end{eqnarray}
Popular regularizers $\Omega(\w)$ include $\Omega(\w) = \lambda\|\w\|_{2}^2$, and $\Omega(\w) = \lambda\|\w\|_{1}$.

\subsection{Auxiliary Image Regularizer}
\label{sec:reg_cnn}

In this work, beyond imposing \emph{a prior} structure of the model $ \mathbf{w} $, we propose a novel regularizer to exploit the data structure within $ D $ to handle the sample noise, which is defined as:
\begin{equation*}
\Omega_{\mathrm{aux}} (\mathbf{w};D) = \|F\mathbf{w}\|_g,
\end{equation*}
where $\|\cdot \|_g$ denotes the group norm defined as $ \|\mathbf{v} \|_g := \sum_{j \in \mathcal{G}} \lambda_j\|\mathbf{v}_j\|_{2}$. In the group norm, $\mathcal{G}$ is an index set of all groups/partitions within the vector $ \mathbf{v} $, $(\lambda_j)_{j\in \mathcal{G}}$ are some positive weights, and $\mathbf{v}_j$ denotes the sub-vector indexed by group $ j $. The norm $\|\mathbf{v} \|_g$ induces a group sparsity that encourages the coefficients outside a small number of groups to be zero. 

We do not impose the group norm on the parameter $ \mathbf{w} $ directly. Instead, we encourage the unlabeled response of the deep model on the learned representation of image data set $ D $ to be group-wisely sparse, such that only a small number of images will contribute to learning of the model, while other non-relevant images are filtered out. These image representations are extracted from the model pre-trained on the well-labeled data set $D^\prime$. Therefore, the regularization draws information from features trained on available data $ D^\prime$.

Each image is a ``group" of active features. The regularizer enforces the features of the ``good" and ``stable" images to be used for model learning, while noisy additional activations will be disregarded. The active images are the ones that are categorized well in the feature space obtained from the deep model. For a new image, the subset of active features close to those stable images in the learned manifold will be weighted most highly (hence ``neighbor regularization").

Towards this target, the matrix $F$ is constructed as $ F^\top = [\mathbf{X}_1, \mathbf{X}_2, \cdots, \mathbf{X}_n] $ and $ (F\w) = [{\mathbf{X}_1} \w, {\mathbf{X}_2} \w, \cdots, {\mathbf{X}_n} \w]^\top$. Here, $\mathbf{X}_i$ is a diagonal matrix consisting of the  features (such as the outputs of the fc-7 layer) of image $i$ representing the $i$-th group in the group norm regularization setting. The group norm enforces the resulting response vector $F\w$ to be sparse, namely only a small number of images are active. These active images are auxiliary examples for learning that are automatically identified by the model. They contribute additional information to model learning. In multi-class classification, $F\w$ is a matrix and the group norm regularizer is defined as the sum of all group sparsity norms of each column. 


Our proposed auxiliary image regularized (AIR) model is then defined as:
\begin{equation*}
\w = \arg\min L(\w;D) + \Omega_{\mathrm{aux}}(\w;D).
\end{equation*}

Our goal is to classify an independent target data set $D$ that has corrupted labels. 

A deep Convolutional Neural Network (CNN) is parameterized layer-wisely by $ \w^{(1)},\ldots, \w^{(\ell)} $, and in this work we only impose the auxiliary regularizer on the top (last) layer of the CNN. The objective function we are going to work with for training a deep CNN model is
\begin{equation*}
L( \w^{(1)},\ldots, \w^{(\ell)};D) + \Omega_{\mathrm{aux}}(\w^{(\ell)};D).
\end{equation*}
A popular loss function used for image classification with CNNs is the following cross-entropy loss with softmax: 
\begin{eqnarray}
\label{eq.1}
L(\mathbf{w};D) = -\sum_{i=1}^{n} \sum_{j=1}^{|\mathcal{C}|} \mathds{1} \{y_i=j\}\log \left\{\frac{\exp \{{\w_j^{(\ell)}}^\top \x_i \} } {\sum_{h=1}^{|\mathcal{C}|} \exp \{{\w_h^{(\ell)}}^\top \x_i\}}\right\},
\end{eqnarray}
with the output feature $ \mathbf{x}_i $ being a function of $ \w^{(1)},\ldots, \w^{(\ell-1)} $ and the raw input image. Here, $\mathds{1} $ is an indicator function.

\section{Optimization with Stochastic ADMM}
\label{sec:opt}
The standard backprogapation technique via stochastic gradient descent cannot handle the non-smooth regularization term $\Omega_{\mathrm{aux}}$ well and presents slow convergence rate. We here demonstrate how the loss function in Section~\ref{sec:reg_cnn}  can be optimized via  ADMM. In this section, we use $\w$ to denote the parameter of the top layer $\w^{(\ell)}$. 
 
After introducing an auxiliary variable $ \mathbf{v} = F\mathbf{w} $, our  optimization problem becomes:
\begin{eqnarray}
\label{eq1}
\min_{\mathbf{v},\w} && L(\w;\x, y)+ \lambda_{1} \|\mathbf{w}\|_{2}^2 +  \sum_{j \in \mathcal{G} } \lambda_{g} \|\mathbf{v}_g\|_{2},  \nonumber \\
\text{s.t. } && \mathbf{v}=F\w,
\end{eqnarray}

The matrix $F \in \mathbb{R}^{np \times p}$ is a sparse matrix with only one non-zero entry in each row, which corresponds to an entry in the feature vector, and the first $p$ rows of $F$ correspond to the features of the first sample, indicated as $\mathbf{X}_1$ in Section~\ref{sec:reg_cnn}. In other words, the matrix $F$ gives a weight to the parameter value $w_i$ ($1 \leq i \leq p$) within each group that is equal to the $i$-th entry in the feature vector. Feature values indicate the activity of corresponding visual feature of an image in the network model. Each term in the group regularizer is normalized by the size of the group through weights $\lambda_g$ such that all of the groups have the same overall effect in the regularizer.

The approximated Augmented Lagrangian \citep{ouyang2013stochastic} for the objective function in Eq.~\eqref{eq1} with $\mathbf{u}$ being the Lagrange variable is:
\begin{eqnarray}
\min_{\mathbf{v},\w} && L(\w_k;\x, y)  + \langle \mathbf{g}_k,\w \rangle + \lambda_1 \|\mathbf{w}\|_{2}^2 +\sum_{j \in \mathcal{G} } \lambda_g \|\mathbf{v}_g\|_{2} \nonumber
 \\
&&+\langle \mathbf{u}, \mathbf{v} -F\w \rangle +\frac{\rho_k}{2} \|\mathbf{v}-F\w\|_{2}^2 + \frac{1}{2\eta_{k+1}} \| \mathbf{w}-\mathbf{w}_k\|_{2},
\end{eqnarray}
where $L(\w;\x,y)$ is replaced with its first order approximation at $\w_{k}$, with $k$ indicating the number of iteration. Here, $\mathbf{g}_k$ is the gradient of $L(\w;\x, y)$ at the $k$-th iteration over a mini-batch of training samples. We set $\eta_k$ equal to $2/(k+2)$ according to \citet{azadi2014sadmm}. Then applying Stochastic Alternating Direction Method of Multipliers (SADMM) \citep{ouyang2013stochastic} followed by a non-uniform averaging step to give higher weights to the recent updates, inspired by \citet{azadi2014sadmm}, gives the following alternative updates of the variables $\w,\mathbf{v},\mathbf{u}$:
\begin{eqnarray}
\w_{k+1}&=& \argmin_{\w}  L(\w_k;\x,y) + \langle \mathbf{g}_k,{\w} \rangle + \lambda_1 \|\mathbf{w}\|_{2}^2 \nonumber \\
&+& \frac{\rho_k}{2} \left\|F\w-(\mathbf{v}_k+{\mathbf{u}_k}/{\rho_k})\right \|_{2}^2 + \frac{1}{2\eta_{k+1}} \| \mathbf{w}-\mathbf{w}_k\|_{2}\label{eq.w} \nonumber\\
\mathbf{v}_{k+1}&=&\argmin_{\mathbf{v}} \sum_{j \in \mathcal{G}} \lambda_g\|\mathbf{v}_j\|_{2} + \frac{\rho_k}{2} \left\|\mathbf{v} -\left(F\w_{k+1}-{\mathbf{u}_k}/{\rho_k}\right) \right\|_{2}^2 \label{eq.v} \\	
\mathbf{u}_{k+1}&=&\mathbf{u}+\rho_k(\mathbf{v}_{k+1}-F\w_{k+1})\nonumber\\
\rho_{k+1}&=&\beta \rho_{k} \nonumber
\label{eq.u}
\end{eqnarray}
We use an adaptive $\rho$ by increasing its value by a factor of $\beta$ in each iteration up to a fixed maximum value $\rho_{\text{max}}$. $\mathbf{w}$ can be updated in a closed form solution as:

\begin{eqnarray*}
&&\w_{k+1}= (\rho_k F^\top F + \frac{I}{\eta_{k+1}})^{-1} (-\mathbf{g}_k - \frac{\lambda_1}{2} \w_k+\rho_k F^\top \mathbf{v}_k + F^\top \mathbf{u}_k + \frac{1}{\eta_{k+1}}\w_k ) \label{eq.w2}
\end{eqnarray*}

 The update of $\mathbf{v}$ has a closed-form solution as the proximal operator for the $\ell_{1/2}$-norm (group soft-thresholding)~\citep{bach2011convex}:
 \begin{eqnarray}
\mathbf{v}_g^{k+1}=  \text{prox}_{2, \lambda_{g}/\rho_k}  \left(F_g\w_k-{\mathbf{u}^k_g}/{\rho}\right),
 \end{eqnarray}
where  soft-thresholding operator for group norm is defined as
\begin{equation*}
 \quad
    \text{prox}_{2,\alpha} (z) = \begin{cases}
    	0, & \text{if}\ \|z\|_{2} \le \alpha \\
    	\frac{\|z\|_{2}-\alpha}{\|z\|_{2}} z , & \text{otherwise.}
    \end{cases}
\end{equation*}

Since the update of $\mathbf{v}_i$ is independent of the update of any other $\mathbf{v}_j$ when $i\neq j, i,j \in \mathcal{G}$, all $\mathbf{v}_i$ can be updated in parallel to further reduce computational cost.

Thereafter, we apply a non-uniform averaging step on $w$ as:

\begin{equation*}
\bar{\w}_{k} = (1-\theta_k) \bar{\w}_{k-1} + \theta_k \w_k,
\end{equation*}

where $\theta_k = 2/(k+2)$; similar updates apply for $\bar{\mathbf{v}}_k$ and $\bar{\mathbf{u}}_k$.


\section{Experiments on Deep CNN Model}
\label{sec:expts}

In this section, we explore the robustness added to the classifier by the proposed auxiliary regularizer and evaluate 
the performance of the regularized CNN model proposed in Section \ref{sec:reg_cnn} on different benchmark data sets.
First, we examine the effect of the auxiliary regularizer when noisy labels are added to clean data sets manually. Second, we investigate its influence on the robustness of the model trained on a freely-available user-tagged data set.

In all of our experiments, we use the AlexNet CNN model pre-trained on the ISLVRC2012 data set \citep{jia2014caffe} and fine-tune its last layer on the target data set $D$. Furthermore, we set $\lambda_1$ in Eq.~\eqref{eq1} to a very small number, $\lambda_{g}$ equal to $10$ over the length of each feature vector, and fine-tune the other set of hyper-parameters (batch size, $\beta \in \{1.1, 1.3, 1.5\}$  , and $\rho_{\text{max}}$) in the SADMM updates of Eq.~\eqref{eq.v} on the cross-validation set for each experiment. The initial value for $\rho$ is $10$ in all experiments. We cross-validate the regularization parameter $C$ for our SVM baseline from the set of $\{1,10,100\}$. We define our loss function to be a softmax and apply the AIR regularizer on the top layer in the CNN model.

\subsection{Experiments with Synthetic Noisy Labels}
\label{sec-synth}

First, we conduct image classification on a subset of the ImageNet7k data set \citep{deng2010does}. We use a pre-trained AlexNet CNN model and fine-tune its last layer on randomly  selected $ 50 $ classes from ImageNet7k data set as the leaf categories of  animal each of which contains $100$ samples. We randomly flipped half of the labels among all 50 categories. This is exactly the problem described in Section \ref{sec:problem}. 

We perform similar experiment on the MNIST data with 10 categories of handwritten digits containing $7 \times 10^4$ samples in total. We use a confusion matrix $\mathbf{Q}$ to define the distribution of noisy labels among the 10 categories. We followed the same settings as in \citet{sainbayar2015} to determine the probability of changing label $i$ to label $j$ for all $i,j \in [1,10]$ as $q_{ij}$. Different levels of noise are applied by setting the diagonal values of the $10\times 10$ matrix $\mathbf{Q}$ equal to the noise level and normalizing the distribution accordingly. 

We empirically investigate the robustness gained from applying the proposed AIR regularize compared  to a linear SVM classifier as well as Robust Logistic Regression (RoLR) \citep{feng2014robust} which assumes a constant fraction of outliers \textendash{} The SVM and RoLR classifiers are used on the last layer of a similar network. The results, shown in Table \ref{tab-synth1}, demonstrate that AIR offers significant improvements over the performance of deep CNN plus SVM or RoLR. On the ImageNet7k data, the performance gain is as large as $8\%$ compared to the SVM.

\begin{table}[t]
\caption{Comparison of classification accuracy achieved by the deep models with different regularizers: ``ft-last-AIR" as our proposed AIR regularizer, ``ft-last-SVM" as SVM with $\ell_2$ regularization, and ``ft-last-RoLR" as the Robust Logistic Regression method, all applied on fine-tuning the last layer. All models were pre-trained on the ILSVRC2012 data set and fine tuned on the target data set.}
\label{tab-synth1}
\begin{center}
\begin{tabular}{lll}
\multicolumn{1}{c}{\bf Regularizer}  &\multicolumn{1}{c}{\bf data set}
\\ \hline 
						& {\bf ImageNet7k} & {\bf MNIST}\\
\hline 
AlexNet+ft-last-AIR         &$\bf{61.5\%}$  &$\bf{92.37\%}$ \\
AlexNet+ft-last-SVM           &$53.44\%$  &$90.93\%$ \\
AlexNet+ft-last-RoLR			& $58.71\%$ &$90.89\%$\\
\hline
\end{tabular}
\end{center}
\end{table}

\begin{figure}[t!]
	\centering
	\includegraphics[width=\textwidth]{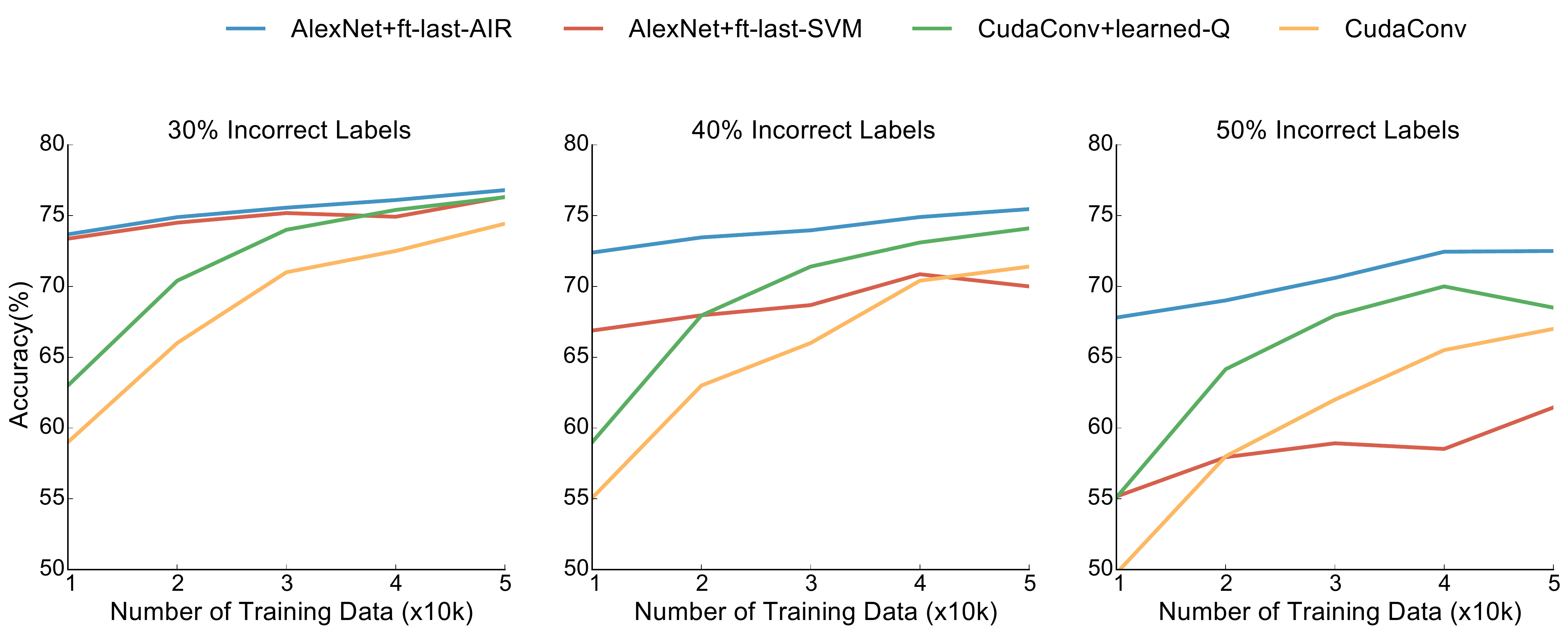}
\caption{Comparison of different deep models on CIFAR-10: ``ft-last-AIR" as our proposed AIR regularizer, and ``ft-last-SVM" as SVM with $\ell_2$ regularization applied on the last layer of Alexnet. ``CudaConv+learned-Q" indicates the end-to-end model proposed by \citet{sainbayar2015} where they add a noise layer to the ``CudaConv" model.}   
	\label{fig.cifar10}
\end{figure}

We also trained the deep CNN with AIR on the CIFAR-10 data which contains 10 different categories each with $5000$ training samples. We used the same confusion matrix $\mathbf{Q}$ as explained above and exactly the same as the matrix defined in \citet{sainbayar2015} to randomly corrupt the labels. Since the size of images in CIFAR-10 and MNIST data sets is small ($32 \times 32$ and $28 \times 28$ respectively), we re-sized images to $256 \times 256$, and then cropped them to $227 \times 227$ and thereafter extracted their fc7 features from the pre-trained model explained above. We tested the robustness of the model on different batches of CIFAR-10 with various noise levels. In Figure \ref{fig.cifar10}, we compare the accuracy of the AIR model to that of an SVM with $\ell_2$ regularization, \citet{sainbayar2015}'s deep model that adds a noise layer to the CudaConv network, and the CudaConv network. CudaConv refers to the network with three convolutional layers with similar model architecture and hyper-parameter settings used in \citet{sainbayar2015} given by \citet{Krizhevsky-cudaconv}.

 Figure~\ref{fig.cifar10} illustrates that the SVM model suffers as the number of incorrect labels grows, whereas AIR remains robust even with large amounts of corruption. Moreover, the performance of both CudaConv+learned-Q and CudaConv depends heavily on the number of training samples, while we can reach the same classification accuracy with significantly fewer number of data points and the same noise level.   

To show the benefit of stochastic ADMM in solving our proposed regularized optimization problem in comparison with stochastic gradient descent (SGD), we re-run the same experiment on CIFAR-10 data set but train the model with SGD in the presence of $50\%$ label noise. This experiment results in an accuracy of $66\%$ compared to $72\%$ accuracy achieved by stochastic ADMM following the same settings discussed in section~\ref{sec:opt}.

We also trained AlexNet with $\ell_2$ regularizer in an end-to-end scheme on CIFAR-10 with $50\%$ incorrect labels where the weights of the last layer are initialized with the learned weights from AlexNet+ft-last-SVM and AlexNet+ft-last-AIR. Classification accuracy obtained from these different initializations is $78\%$ and $79\%$, respectively. It shows that even implementing the proposed AIR regularizer in the last layer will improve the accuracy of the whole deep model that is trained end-to-end. 

\begin{figure}[t]
	\centering
	\includegraphics[width=\textwidth]{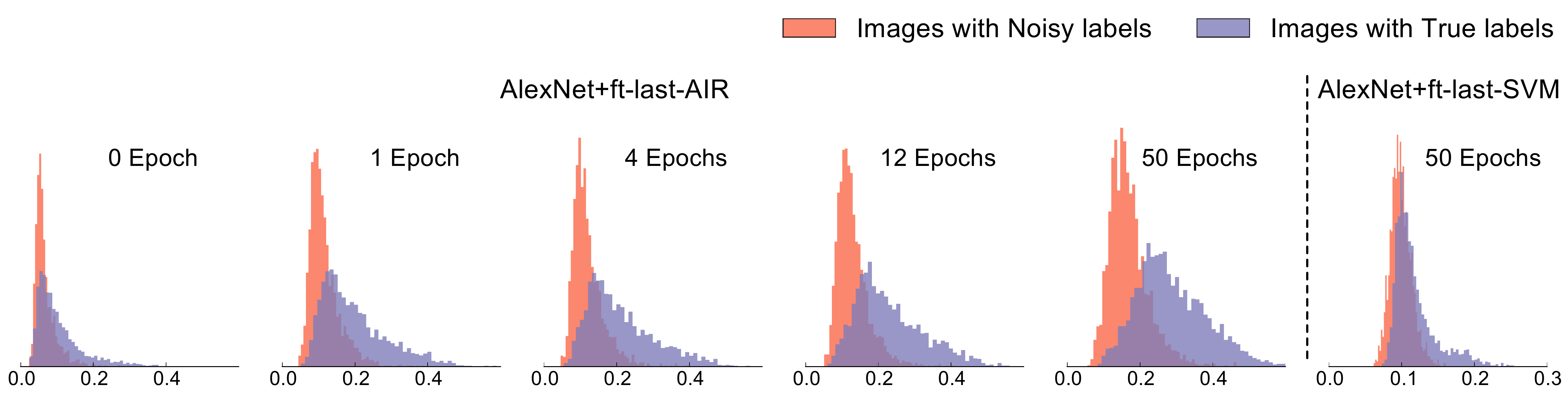}
\caption{Distribution of activations given to true and noisy images along different learning iterations. Images belong the Imagenet7k experiment. Distributions on the left hand side of the plot refer to AIR and the one in the right is related to SVM.}   
	\label{fig.dist_v}
\end{figure}

The AIR regularizer automatically incorporates inherent structure in the image data by forcing the weights of noisy groups to decrease toward zero and giving higher learning weights to the stable groups. We illustrate this characteristic of AIR in Figure~\ref{fig.dist_v} by comparing the distribution of activations of noisy-labeled images and the distribution of the activations of clean images in the Imagenet7k experiment along different learning iterations. Here ``activation" refers to the $\ell_2$-norm of the weights associated with each of the groups, \emph{i.e.}, $\| \mathbf{v}_g\|_{2}$ in Eq.~\eqref{eq1}. The distributions of these two sets of activations highly overlap in the first iteration and they gradually get more and more distinct from each other revealing the ability of the auxiliary regularizer in finding the images with noisy annotations. We manually compute the same activation scores per image for SVM by setting $\mathbf{v}_g^{\text{SVM}} =\mathbf{X}_g \w^{\text{SVM}}$ and compare the corresponding distribution in the right plot of Figure \ref{fig.dist_v}. Activation scores for images with both clean and noisy labels, learned from SVM, overlap notably after equal number of epochs of training with AIR.
 
\begin{figure}[t]
    \centering
    \begin{subfigure}[c]{0.47\textwidth}
        \includegraphics[width=\textwidth]{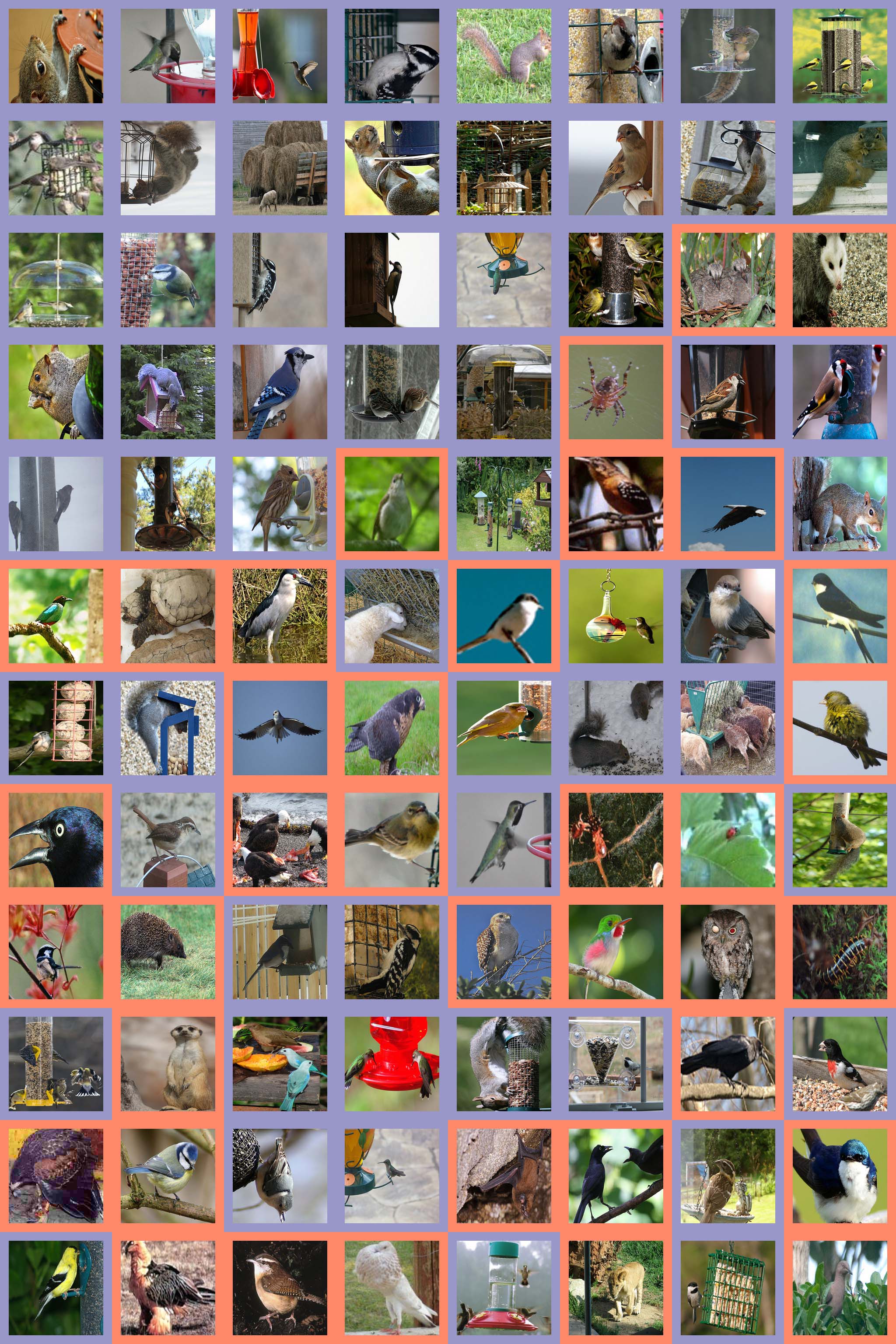}
        \caption{Auxiliary Image Regularizer (AIR)}
        \label{grpnrm}
    \end{subfigure}
    \hfill
    \begin{subfigure}[c]{0.47\textwidth}
        \includegraphics[width=\textwidth]{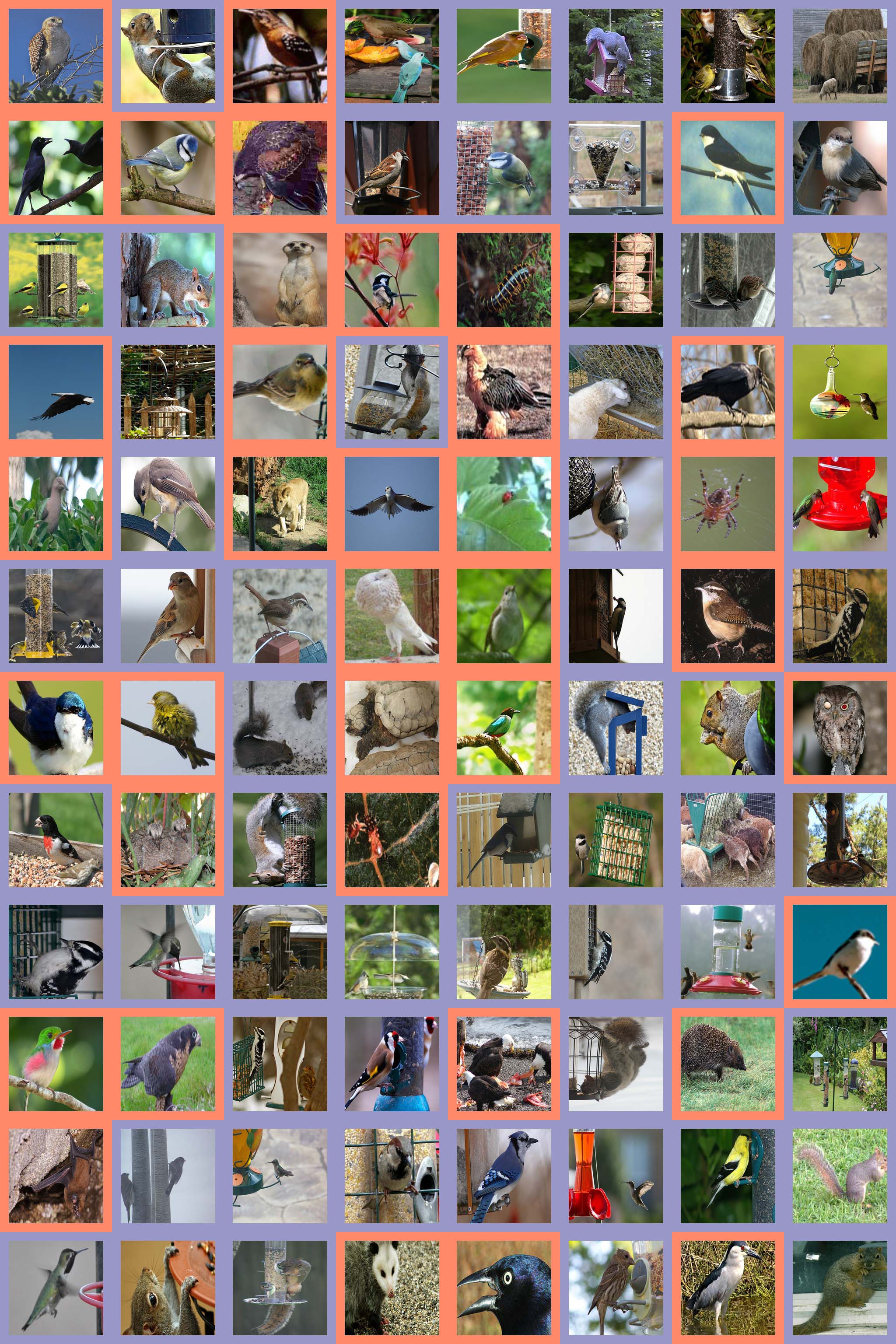}
        \caption{$\ell_2$-Regularized SVM}
        \label{svm}
    \end{subfigure}
    \caption{Images with clean labels have blue borders and images with noisy labels are surrounded by red borders. Images are ranked based on their activities found by AIR (right) or SVM (left) from the top row to the bottom. Images are from a category with synset ID ``n01317916" from the $50$ animal categories in the ImageNet7k data set.}
    \label{fig.corrupted_best_worst}
\end{figure}

\subsection{Experiments with Real Noisy Labels}
\label{sec:real-noise}
In this section, we examine the performance of deep image classifier on images with user-supplied tags from publicly available multi-label NUS-WIDE-LITE data set \citep{nus-wide-civr09} as a subset of large Flickr data set.
This data set contains 81 different tags with the total number of 55615 samples divided into two equal-sized train and test sets. After ignoring the subset of training samples which are not annotated by any tags, we have $20033$ samples in the training set. We train the classifier on the user-tagged images of training data and evaluate the performance on the test set with ground-truth labels. 

\begin{figure}[t!]
	\centering
	\includegraphics[width=0.6\textwidth]{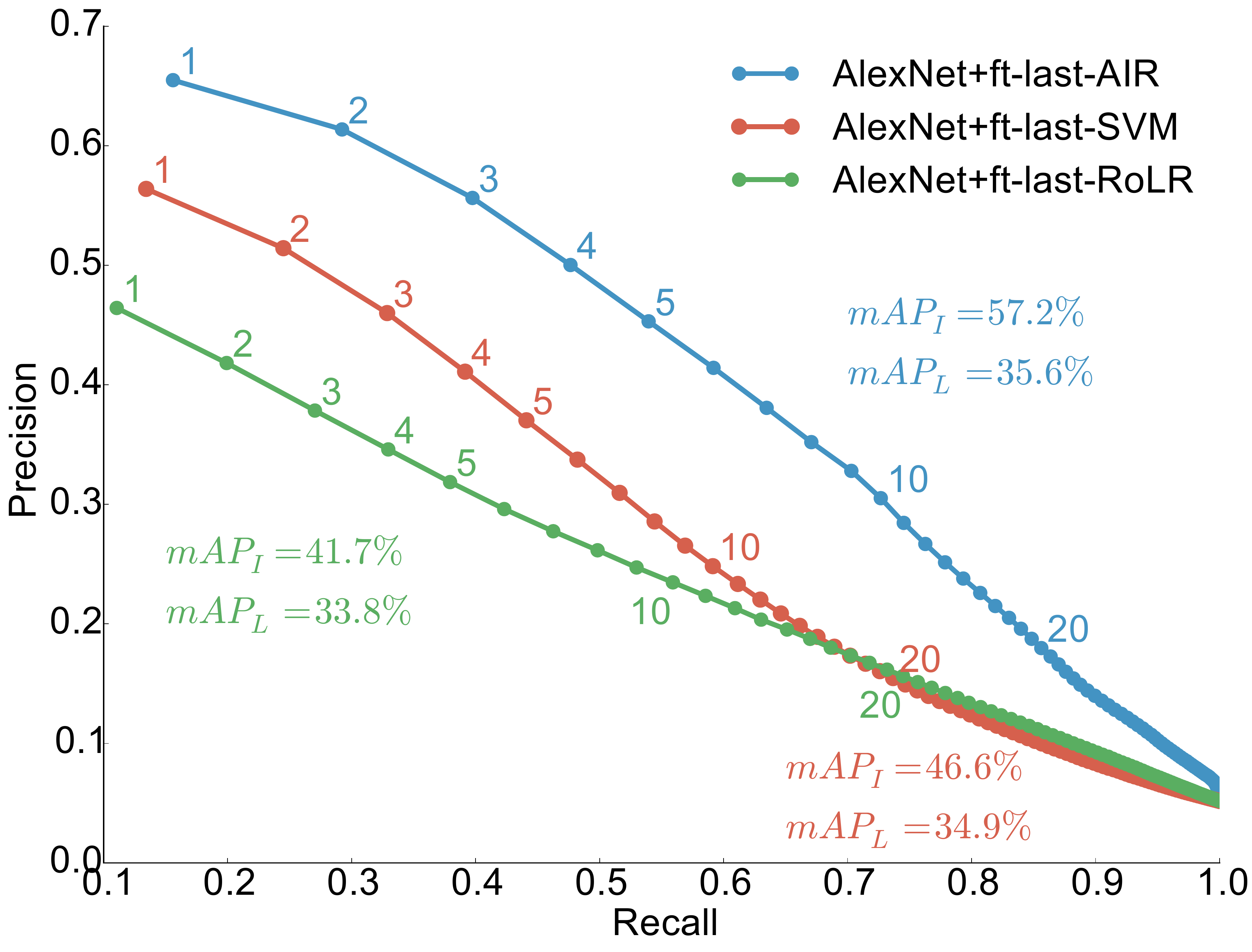}
\caption{Comparison of overall precision and recall for different methods on NUS-WIDE-LITE data set. Numbers on each plot represent the number of top labels used to calculate the corresponding precision and recall. $mAP_L$ and $mAP_I$ for each method has also been reported beside its corresponding curve.}   
	\label{fig-nus}
\end{figure}

\begin{figure}[t!]
	\centering
	\includegraphics[width=\textwidth]{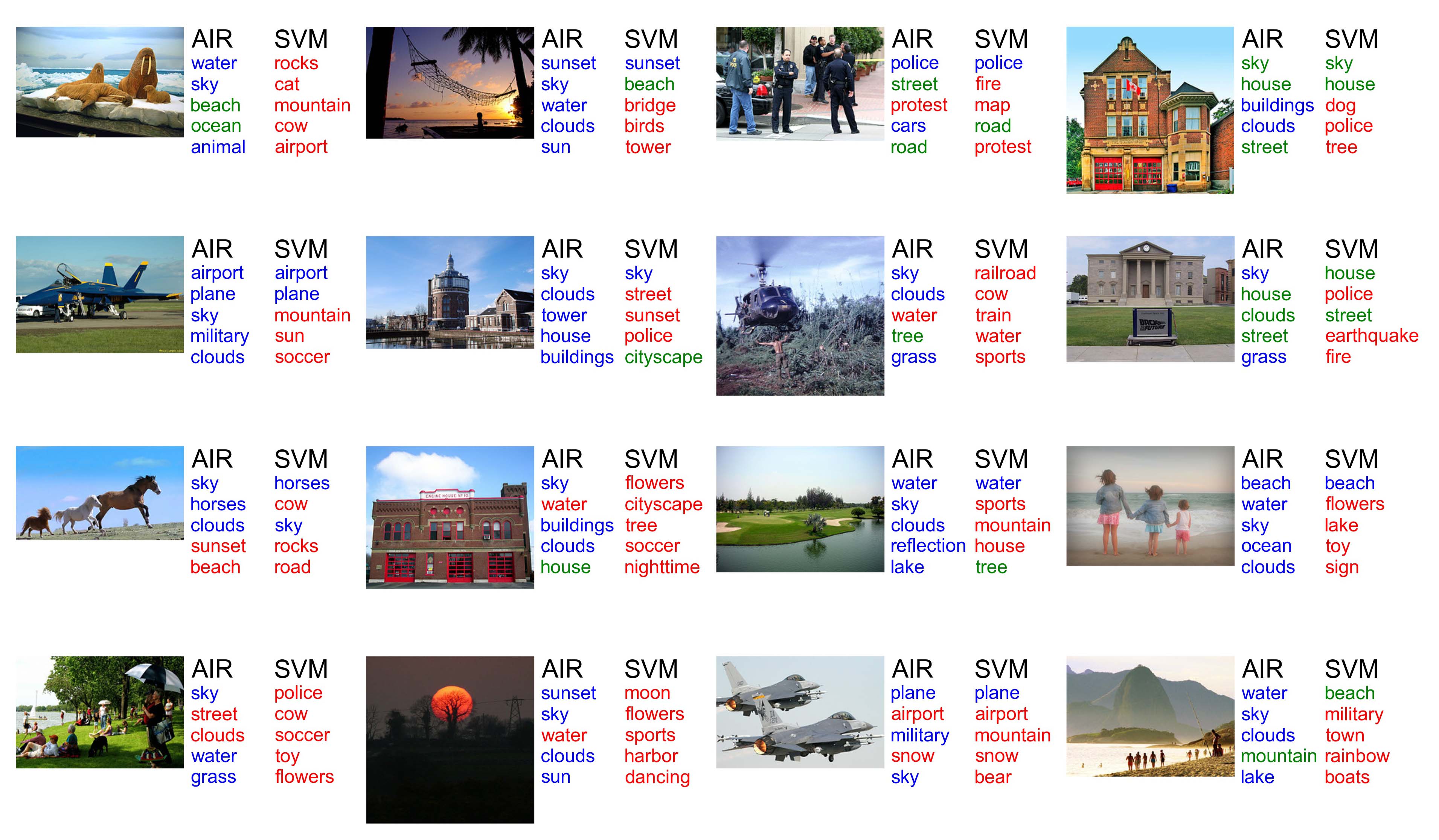}
\caption{Instance results for top-5 predictions for a few examples of NUS-WIDE-LITE data set by AIR and SVM. We illustrate true predictions (compared with ground-truth annotations) in blue, predictions non-overlapping with the ground-truth labels in red, and predictions looking as reasonable annotations but not in the ground-truth labels in green.}
	\label{fig-flickr_predictions}
\end{figure}

We followed the same experimental settings as explained in Section \ref{sec:expts}. We compare the performance of deep model with different classifiers applied to the last layer of AlexNet. In Figure~\ref{fig-nus}, we plot averaged-per-image precision and recall values when we assign $n$ highest-ranked predictions to each image. As a secondary metric, we compare AIR's performance with the baselines in terms of Mean Average Precision ($mAP$) \citep{Li2015semantic} which does not depend on the top rankings, but on the full rankings for each image. We use $mAP_L$ to measure the quality of image ranking per label and $mAP_I$ to measure the quality of tag ranking per image. Robustness to noisy user-tags obtained from the auxiliary regularizer is significant as shown in Figure \ref{fig-nus}. A few sample images and their top-5 predictions by both AIR and SVM regularizers are represented in Figure~\ref{fig-flickr_predictions}.

\subsection{Visualization of Selected Auxiliary Images}
Finally, we try to understand whether the model indeed has the expected ability of retrieving informative images during the training process. To this end, we visualize the automatically selected auxiliary images in Figure~\ref{fig.corrupted_best_worst}, left plot. The figure refers to the experiments on the Imagenet7k data set with noisy labels with the same settings explained in Section~\ref{sec-synth}, and displays the images whose corresponding groups were active (selected during the optimization), and also the images that were filtered out (suppressed). The images in this figure are ranked based on their activation scores, expecting clean images to have higher ranks and appearing on the top rows of the plot. Similar to Section \ref{sec-synth}, we rank images based on their activation scores obtained from SVM learned weights on the right hand side of Figure~\ref{fig.corrupted_best_worst}. Indeed, the figure shows that AIR forces the weights of noisy or non-informative images to zero and encourages the model to select clear and informative images in the training procedure much more accurately than SVM. This also explains why our proposed model is robust to the noisy-labeled training examples, as shown in the previous experiments.

 \subsection{Scalability of the Auxiliary Image Regularizer}
To reduce memory requirement for matrix $F$ on large data sets, we can randomly select a small number of groups to be considered in AIR regularizer. When large number of data points are available, randomly ignoring the groups in regularization of the response will not substantially affect the learning process that is influenced by the distribution of informative images in the feature space. To verify this point, we repeat the experiment on CIFAR-10 data set with $50\%$ synthetic noise level but with only $1\%$ of the groups used in the regularization term. This significant memory reduction will only drop the accuracy by $0.5\%$ from $72.3\%$ to $71.8\%$. This experiment shows sampling from the groups does not reduce the final classification accuracy considerably in the case of large data sets but saves the memory cost significantly.

 \section{Summary and Future Work}
We introduced a new regularizer that uses overlapping group norms for deep CNN models to improve image classification accuracy when training labels are noisy. This regularizer is adaptive, in that it automatically incorporates inherent structure in the image data. Our experiments demonstrated that the regularized model performs well for both synthetic and real noisy labels: it leads to a substantial enhancement in performance on the benchmark data sets when compared with standard models. In the future, we will explore the effect of AIR on robustifying the classifier in an end-to-end scheme where the error information from the auxiliary regularizer will back-propagate through the inner layers of deep model to produce robust image representations.
\newpage

\bibliography{newBib}
\bibliographystyle{iclr2016_conference}

\end{document}